\documentclass{article}

\usepackage{arxiv}

\usepackage[utf8]{inputenc} 
\usepackage[T1]{fontenc}    
\usepackage{hyperref}       
\usepackage{url}            
\usepackage{booktabs}       
\usepackage{amsfonts}       
\usepackage{nicefrac}       
\usepackage{microtype}      
\usepackage{lipsum}		
\usepackage{graphicx}
\usepackage[square,numbers]{natbib}
\usepackage{doi}
\usepackage{algorithm2e}

\title{Combating the effects of speed and delays in end-to-end self-driving}

\author{ \href{https://orcid.org/0000-0002-2120-1712}{\includegraphics[scale=0.06]{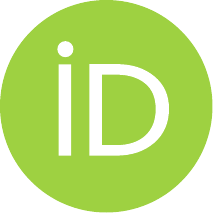}\hspace{1mm} Ardi Tampuu}\thanks{All authors were affiliated with University of Tartu during this research} \\
	\texttt{ardi.tampuu@ut.ee} \\
	\And
	Ilmar Uduste \footnotemark[1]\\
 \AND
        Kristjan Roosild \footnotemark[1]\\ 
}

\renewcommand{\shorttitle}{\textit{arXiv} Template}

\hypersetup{
pdftitle={A template for the arxiv style},
pdfsubject={q-bio.NC, q-bio.QM}, 
pdfauthor={Ardi Tampuu},
pdfkeywords={End-to-end driving, delays, out-of-distribution},
}

\begin{document}
\maketitle

\begin{abstract}
In the behavioral cloning approach to end-to-end driving, a dataset of expert driving is collected and the model is optimized to guess what the expert would do in different situations. Situations are summarized in observations and the outputs are low or mid-level commands (e.g. brake, throttle, and steering; or trajectories). In short, the models learn to match observations at time T to actions recorded at T or as simultaneously as possible. However, when deploying this model to the real world (or to an asynchronous simulation), the action predicted based on observations at time T gets applied at T + $\Delta$ T. In a variety of cases (e.g. low compute environment), $\Delta$ T can be considerable and significantly influence performance. 

In this manuscript, we first demonstrate that driving at two different speeds is effectively two different tasks. Delays partially cause this difference and linearly amplify it. Even without computational delays, actuator delays and slipping due to inertia result in the need to perform actions preemptively when driving fast. This means the function mapping observations to commands becomes different compared to slow driving. We experimentally show that models trained to drive fast cannot perform the seemingly easier task of driving slow and vice-versa. This means that good driving models may be judged to be poor due to testing them at "a safe low speed", i.e. in a task they cannot perform. 

As a separate contribution, we show how to counteract the effect of delays in end-to-end networks by changing the target labels. This is in contrast to the common approaches attempting to minimize the delays, i.e. the cause, not the effect. To exemplify the problems and solutions in the real world, we use 1:10 scale minicars with limited computing power, using behavioral cloning for end-to-end driving. However, we believe some of the ideas discussed are transferable to the wider context of self-driving, to vehicles with more compute power and end-to-mid or modular approaches.
\end{abstract}

\keywords{End-to-end driving \and delays \and out-of-distribution}

\section{Introduction}
The fully end-to-end driving systems convert raw sensory inputs directly into actionable driving commands via a single neural network model \cite{ tampuu2020survey,ly2020learning,huang2020autonomous,chen2023end}. Imitation learning, especially behavioral cloning has been the dominant training paradigm for training such models \cite{pomerleau1989alvinn, codevilla2019exploring,chitta2022transfuser, chen2023end} in the past ten years. The camera feed is often the only input \cite{bojarski2016end, osinski2019simulation, bansal2018chauffeurnet, kendall2019learning,chekroun2023gri} and is fed into the network from the input end. From the output end, the model attempts to predict human-like steering angle, throttle, and brake values, i.e. "what would the human expert do in the given situation". These directly actionable "end" commands can be passed into the drive-by-wire system of the vehicle. Alternatively, end-to-mid approaches would output a desired trajectory \cite{bansal2018chauffeurnet}, cost map \cite{zeng2019end}, or other not directly actionable output representations\cite{hu2021safe}, which need to be further processed by subsequent modules. In the following, we study camera-based fully end-to-end driving, while some of the problems raised and solutions proposed may generalize to a wider variety of approaches.

As a major limitation of behavioral cloning, the predictive task of transforming images to actionable commands that the model is trained to perform differs from the actual task of driving \cite{codevilla2019exploring}. Below we give a list of differences between the tasks of predicting accurately on a static dataset and the task of controlling a vehicle on the road.
\begin{itemize}
    \item \textbf{The type of mistakes that matter differs.} Driving is a sequential decision-making process, whereas behavioral cloning optimizes the model to predict well for individual situations randomly assigned into mini-batches. Temporal distribution of errors matters for the sequential decision-making task \cite{codevilla2018end}, the car can drift off the good trajectory little by little as a result of tiny but correlated errors. In behavioral cloning with usual loss functions and sampling, such consistent but weak biases are penalized minimally. A different way of constructing mini-batches or adding a loss penalizing temporally consistent biases might reduce this difference but is not commonly employed.
    \item \textbf{Incoming data differs.} During deployment, the model outputs control the car and implicitly define the distribution of situations the car ends up in. This distribution may significantly differ from the training set situations resulting from expert driving. As a result, behavioral cloning models that predict well on validation and test sets of human driving may fail to generalize well (to the distribution caused by self-driving) when deployed. This well-known effect is often called distribution shift \cite{tampuu2020survey, ross2011reduction}.
    \item \textbf{Delays differ.} Delays play no role in the predictive behavioral cloning task we optimize when creating a model. The loss does not depend on the time spent to compute the answers. However, if the driving model is deployed in the real world, delays exist between the moment an observation is captured by the camera and the moment the actuators receive the computed response to this observation. Furthermore, actuators may have their own delays. Hence, even if the decision was optimal at the time of observing the world, it may no longer be adequate by the time it gets applied.
    \item \textbf{Frequency differs.} Separate from the problem of delays is the problem of decision frequency. Decision frequency, predictions per second, plays no role in the behavioral cloning task we optimize for. In driving, however, if decisions are made too infrequently, each decision gets applied too long and may "overshoot" its intended effect, resulting in an oscillation around the actual desired path. Furthermore, at low frequency, the model can be simply too late to react to situations if they happen between decision ticks.
\end{itemize} 

As a result of these differences between the tasks, the correlation between the ability to predict labels of the held-out human driving data (referred to in the literature as off-policy evaluation, open-loop evaluation) and the ability to drive (on-policy evaluation, closed-loop evaluation) is widely reported to be low \cite{codevilla2018offline,tampuu2020survey,tampuu2022lidar}. 

In our previous work \cite{tampuu2022lidar}, we noticed that the driving speed was another important difference between the way we collected the data (including data for off-policy evaluation) and the way we deployed the model. In particular, we deployed steering-only models on a real vehicle on narrow rural roads, but
\begin{itemize}
    \item we deployed the models at speeds 0.5 to 0.8 times the speed the human used during data collection in the given road section, as extracted from recordings. Similarly, \cite{kendall2019learning} used speeds only up to 10km/h when deploying the models.
\end{itemize}
    
\begin{figure*}
    \centering
    \includegraphics[width=12cm]{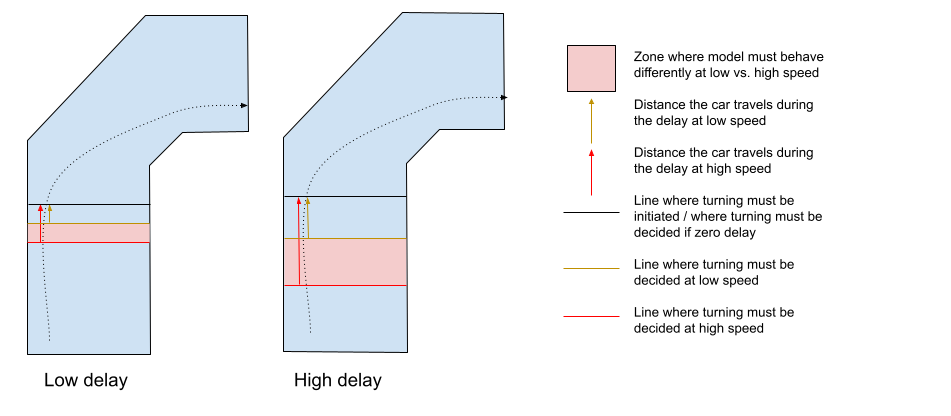}
    \caption{\textbf{Preemptive actions.} When approaching a turn, to stay on the road, the command to turn must be decided preemptively based on the observation from $delay \cdot speed$ meters before the actual turning point. As per the formula, the higher the speed the earlier in terms of location the decision to turn must be taken. Hence, for two different speed values, there exists an area along the road (a set of observations) where the driver or driving model must behave differently. Human drivers also turn preemptively when collecting the data, so this "task difference" is captured in the recorded data. If the delays grow, for example during deployment, the difference between the tasks of slower and faster driving increases. }
    \label{fig:shift}
\end{figure*}

Deploying at a lower speed is safer for the safety driver and is in many ways a smart approach. However, in this work, we argue and demonstrate that deploying at low speeds can cause an out-of-distribution problem for the steering models. We intuitively understand that the on-policy sequential task of driving fast is a different task from driving slow, due to the physics of the world. However, here we argue that the off-policy imitation learning task also differs. The predictive task is defined by the labeled training dataset. During data collection, the expert driver (human, or another model) must account for the chosen speed, so the chosen steering commands, i.e. the labels of the task, differ. In particular, faster driving necessitates preemptive actions to counter inertia, delays in transmitting commands, actuator delays, and computational delays (within the brain if data was collected by a human) while slow driving does not. Importantly, if the input at a given location (e.g. 5 meters before a turn) remains very similar at different speeds, these highly similar inputs need to result in different outputs at different speeds. We refer to this as the \textbf{task shift, a shift in the predictive function from inputs to outputs we attempt to learn, in particular, a shift in the correct outputs for a given input}. The severity of task shift between two speeds increases linearly with delays and with speed difference, as explained in Figure \ref{fig:shift}. 

During deployment, additional computational delays occur, more preemptive actions are needed, decision frequency drops, and in fact the difference between the tasks of driving fast and slow increases. It is intuitive to assign the blame of a model not performing well to the delays. However, we argue that even at trivial computational delays, models trained to drive fast will be inadequate for driving slow and vice versa, because the training data described different tasks in those two cases.

Beyond steering models learned by behavioral cloning, the relevance of the problem of task shift due to speed depends on the model type and how the self-driving task is formulated. In some cases, the imitation learning task is identical at different speeds, but the additional computational delays during deployment re-introduce a higher need for preemptive actions or planning at high speed. We discuss the ramifications of changing the driving speed for some other model types below. 
\begin{itemize}
    \item \textbf{Multi-frame steering models.} Models that receive multiple observations as input (e.g. multi-frame CNN or recurrent nets) have access to implicit information about the traveling speed, contained in the degree of change between frames. However, if such a model is exposed to only a certain speed profile (say 50 km/h at straights and 25 km/h in turns), it will not learn to deduce speed and use it internally. Speed is highly correlated with the scene (a straight or a curve) and adds no further information for solving the predictive off-policy task. If the model does not internally extract and perceive speed, the task shift exists - for very similar inputs, different outputs need to be produced at different speeds. \\
    Moreover, in this work, we show that a sequence of images from a novel speed range causes measurable out-of-distribution-like activation patterns in multi-frame CNN models. So for such models, not only the task (how early and how steep the turning should be) but also the input distribution is shifting.  
    \item \textbf{Joint prediction of speed and steering}. End-to-end models that predict speed and steering jointly must implicitly learn to correct for the task difference. A good model of this type would learn to either predict low speed with corresponding steering values or high speed with matching steering values. However, if the predicted speed is clipped at deployment for safety reasons, the predicted steering value will again become sub-optimal. However, models that produce the full joint probability distribution would allow the selection of the highest probability speed-steering pair in a certain (low) speed range. Similarly, in energy-based models \cite{baliesnyi2023controlling} clipping is not needed - we can select the lowest energy speed-steering pair within a certain speed range. 
    \item \textbf{Reinforcement learning}. If a steering-only policy is learned via reinforcement learning, it will be adequate for only the speeds encountered during training. Learning speed and steering jointly via reinforcement avoids the problem of task shift, but clipping the speed at deployment for safety reasons would reintroduce the issue, as above. 
    \item \textbf{Speed given as input}. Inserting speed as an additional input to a steering-predictor model is another option to avoid task shift - the inputs at different speeds are separable and it is clear which type of steering to produce. However, if the model is trained with only a certain speed range, querying it with an out-of-training-distribution speed value during deployment would naturally cause generalization challenges. From the perspective of the model, this is an input shift (an out-of-distribution input), not a task shift. Energy-based models are a subcategory of this type of model.
    \item \textbf{End-to-mid models}. End-to-mid models outputting trajectory, cost map, or other intermediate representation likely suffer less from the task shift. In a safe speed range, the trajectories an expert driver took when collecting the training dataset likely change only slightly at different speeds. The obstacle maps or cost maps should be identical or similar. In aggressive driving, as an exception, the ideal trajectory does significantly depend on speed. Depending on the way the trajectory is encoded in the output, the model may also control deployment speed. Predicting speed and steering jointly resolves the problem of task shift. Input shift due to speed can still occur, inter-frame similarity decreases when speed increases, and motion blur is amplified, so a trajectory-predicting model might not generalize naturally to a novel speed, even if task shift is minimal. Furthermore, the delays during deployment render the predicted intermediate outputs out-of-date by the time they are produced, for example, self-location along the trajectory should be corrected for the ego-motion that occurred during computation. The amount of this motion depends on speed.
\end{itemize}

Driving at different speeds always constitutes a different task, if not due to delays then due to slip and actuator delays (higher centrifugal force acting against the actuation may increase actuator delay). The presence of computational delays amplifies this difference as all decisions are more spatially belated if we drive fast. Indeed, making decisions late in terms of time does not necessarily lead to crashes (e.g. at speed 0), but being late in terms of space (location on the road) does. Hence delay $\times$ speed, i.e. the spatial belatedness, is an important characteristic of the task the vehicle is performing. According to this aspect, tasks of driving fast in the presence of small delays and driving slow in the presence of higher delays may be very similar unless there is a significant slip or actuator delay.

There are two obvious ways to counteract spatial belatedness - driving slowly and minimizing delays. Here, as our second contribution, we propose a third option - counteracting the effect of computational delays instead of minimizing the delay itself. This can be achieved by conditioning behavioral cloning models to predict commands that are relevant in the future compared to the observation, at the moment when the computation is expected to end. The same method could be applied to trajectory prediction or other end-to-mid prediction targets, but in those cases, it might be easier to correct for the ego-motion during computation in the subsequent control module.

To safely test and demonstrate the effects of speed, computational delays, and our proposed countermeasure, we used 1:10 scale mini-cars equipped with Raspberry Pi 4b and a frontal camera. Our models were trained fully end-to-end with behavioral cloning and controlled only the steering of the vehicle. The training and deployment procedure is very similar to what was done on the real-sized car in our previous work \cite{tampuu2022lidar}. We believe the lessons learned are transferable to the domain of real-sized cars.

The main contributions of the present work are the following:
\begin{enumerate}
    \item We demonstrate that the performance of good driving pipelines may fall apart if deployed at a speed the system was not exposed to during training. We discuss the underlying reasons and the implications for how on-policy testing should be performed. 
    \item We demonstrate to what degree the performance of good driving models suffers due to computational delays. We demonstrate that temporally shifting labels in the training data to take into consideration the delays occurring during deployment allows us to easily alleviate the problem. 
\end{enumerate}

\section{Methods}

\subsection{Experimental design}
We studied the effect of speed and the effect of computational delays as two independent studies based on independently collected data. The work on the track was done as two master theses \cite{roosild2022} and \cite{uduste2022}, independently.

To demonstrate the effect of speed on model performance, we trained models on data collected at certain speeds and measured their generalization ability to a novel speed. We did this for models considering only the latest frame and for multi-frame models considering the past 3 frames. We hypothesized that novel speed causes a performance drop in the ability to predict the labels in the off-policy behavioral cloning task, as well as in the on-policy driving task of driving on the track. 

In the study on counteracting the effect of computational delays, the dataset was collected at high speeds. High speeds were preferred as speed amplifies the effect of delays and makes the results more evident. This collected dataset was transformed into a variety of training sets by shifting the steering values (labels) by one or multiple positions back or forward in time. With 20Hz recordings, every position shift corresponded to 50ms. We trained behavior cloning models on these differently label-shifted data sets, i.e. we created models that predict optimal commands for the past, present, or future. These models were deployed on the track in the presence of different amounts of computational delays and their driving performance was measured. We hypothesized that models predicting future commands are able to drive in the presence of higher delays.

\subsection{Training paradigm}
Based on recordings of expert driving, consisting of camera frames and steering values, we follow the behavioral cloning paradigm \cite{argall2009survey}. This consists of optimizing a neural network to predict the expert-produced steering values based on the training set images.

\subsection{Hardware setup and data collection}
\begin{figure}
    \centering
    \includegraphics[width=8.5cm]{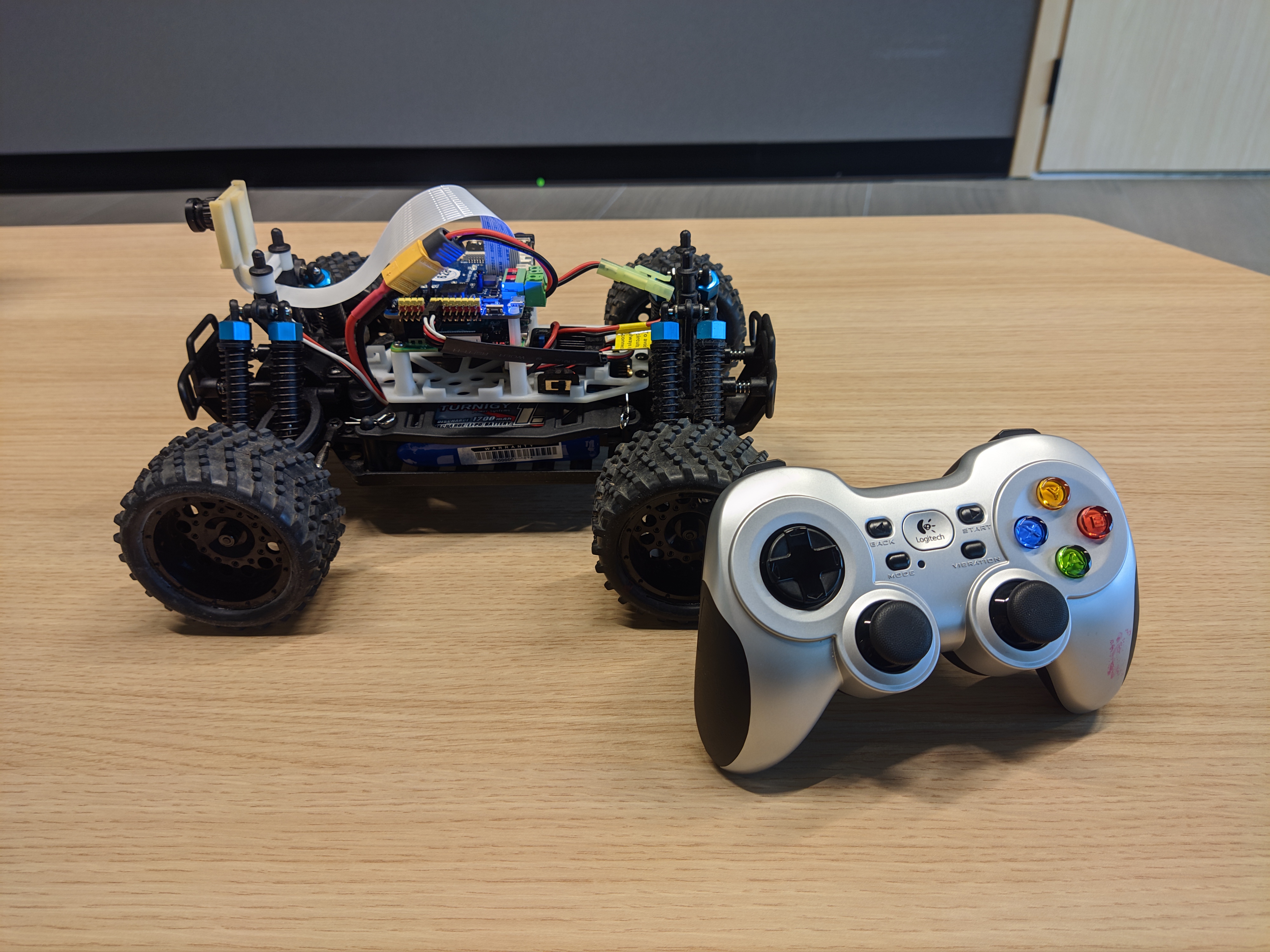}
    \caption{\textbf{Minicar.} The 1:10 scale Donkey Car S1 used in this work, alongside the F710 controller used for human driving data collection.}
    \label{fig:donkey}
\end{figure}

This work is performed using Donkey Car open-source software \footnote{https://docs.donkeycar.com/} deployed on the 1:10 scale Donkey Car S1 platform \footnote{https://www.robocarstore.com/products/donkey-car-starter-kit}, equipped with MM1 control board, Raspberry Pi 4b, Raspberry Pi wide-angle camera (160 degrees), and two servomotors (for steering and throttle). The vehicle can be seen in Figure \ref{fig:donkey}, and is referred to as the mini-car in the remainder of the manuscript. 

\begin{figure}
    \centering
    \includegraphics[width=8.5cm]{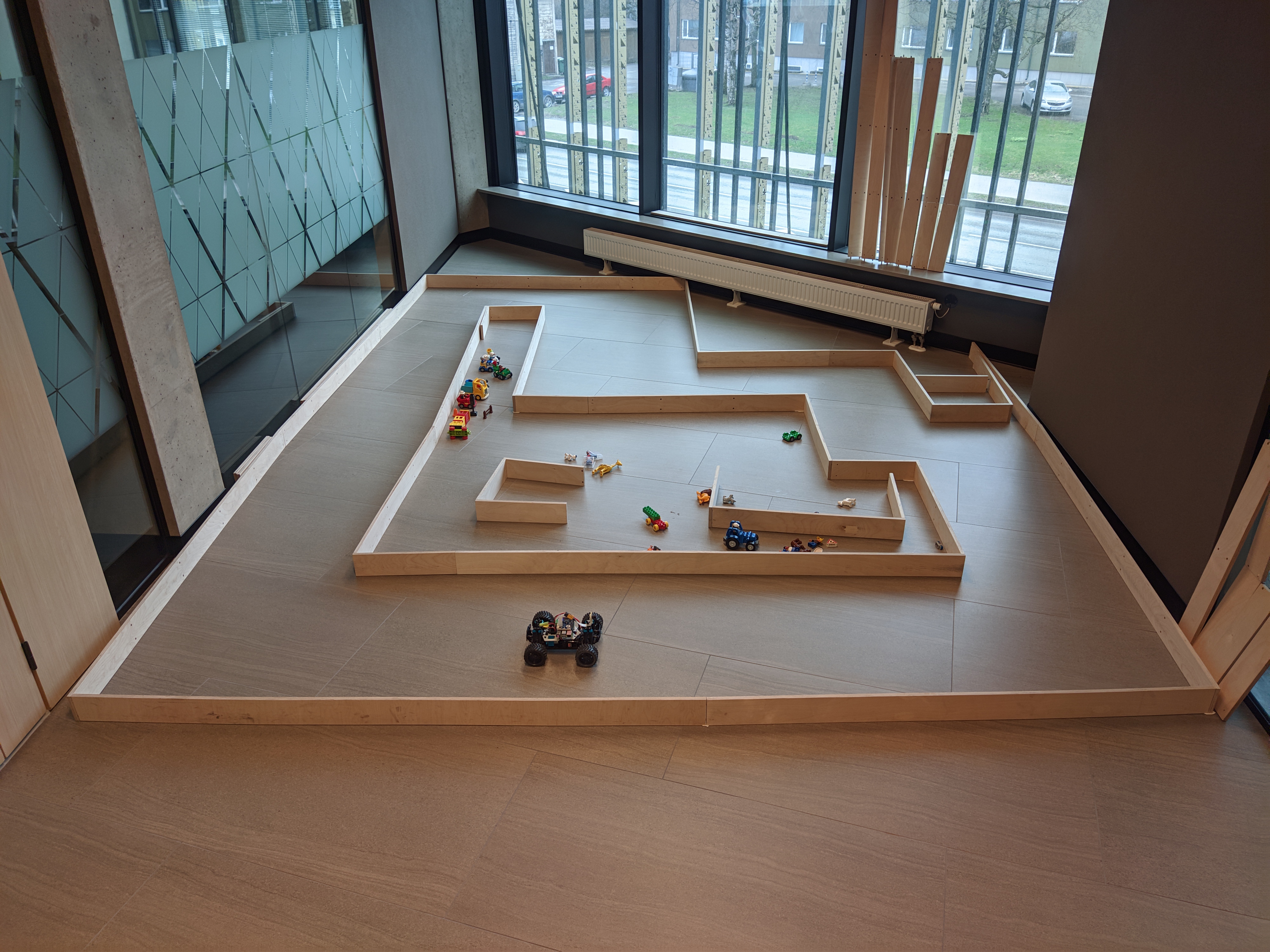}
    \caption{\textbf{The track for data collection and on-policy evaluation.} Driving was done counterclockwise, resulting in five left turns and one right turn. The track width is between 70 and 80 cm. The track length is 17 meters.}
    \label{fig:track}
\end{figure}

Correct assembly and calibration result in a car able to drive very fast (needs to be limited by configuration files so as not to break the hardware) and able to turn with a below 140cm turning diameter (as measured by the exterior wheel). Such a turning radius is approximately equivalent to a minivan's turning diameter (13-15m) and slightly higher than that of smaller consumer cars (10-12m). We deploy the car on a track with 60-80cm of width (see Figure \ref{fig:track}). 

During training data collection, the throttle value can be set to different fixed values by the human operator. The speed resulting from a given throttle value is not stable, as it depends on the battery level and on the heating up of the servomotors. There are no encoders or other means on Donkey Car S1 to measure and assure constant speed. As a result, we do not define driving speed via throttle value but via actual achieved lap times. During data collection, the operator steadily increases throttle value to achieve stable lap times as the car heats and batteries drain. During the on-policy evaluation, the throttle value is also continuously tweaked to achieve lap times in the desired value range.

During data collection, the steering is controlled by the researcher using a Logitech F710 gamepad or, for the study of speed, by a competent self-driving model running on board the mini-car (explained below). The delay between giving a command on the gamepad and the car reacting is not perceivable if no friction with the ground is present (car lifted in the air); hence we conclude the messaging delay is minimal. However, latency in actions (actuator delays) becomes perceivable when driving, likely due to inertia and due to the friction of wheels with the ground. When driving fast, the human driver will naturally compensate for the actuator delays and give commands early, so they would take effect on time. 

In all our driving experiments, all computation happens on board, in the Raspberry Pi 4b device. This holds for both data collection done with the help of a competent "teacher" model and for model evaluations. The duration of different computations is measured by a built-in function of the Donkey Car software. Model training took place on CPUs in laptops or in Google Colab with GPU access and took up to a few hours per model.

\subsubsection{Quality of driving data}
Importantly, it is not easy to yield intermediate valued commands on the Logitech F710 gamepad's stick button, resulting mostly in full-left, full-right, or straight states. At lower speeds, an experienced human operator can achieve a wider distribution of action values, but at high speeds, actions are rushed and mostly extreme. Indeed, higher speeds result in higher inertia and might need steeper turning, but this effect may be exaggerated in the data due to human reaction time and precision of movement.

As an alternative, we train a first-generation self-steering model, a teacher agent capable of driving at various speeds. This teacher model uses the same architecture and training setup as single-frame models in the study of speed (see below) but is exposed to a variety of speeds during training. We then use this model to collect data at two different (slow and fast) speeds. This way the training data at different speeds is still collected by the same expert agent, but this agent is less restricted by physical abilities. While this makes data collection easier, it also alleviates the risk that our conclusions result from low data quality rather than the causes we wish to put in evidence. All results in the section studying the effect of speed are based on models trained on data collected by the teacher model.

The study on counteracting the effect of delays was performed independently and considered training data only in the fast speed range. The data in this second study was collected separately and by a very proficient human expert, not by the teacher agent.

\subsection{Data preparation}
The Donkey Car software allows cleaning the data from unwanted behaviors (e.g. expert driving into the wall) in two ways. First, during data collection, one can delete a certain amount of past frames (by default last 5 seconds) and the corresponding commands from the recording. Secondly, the software allows you to watch through the recordings, and mark and delete unwanted sections.

The cleaned (infractions removed, speed in the designated range) data sets were prepared to demonstrate the two effects:
\begin{itemize}
    \item In the study of speed, we collect recordings at low speeds and at high speeds. In the former, the lap of 17 meters in length is, on average, completed in 24.25$\pm$1.9 seconds. The fast speed corresponds to an average 14.85$\pm$0.8 second lap time. After cleaning, the "fast dataset" size is 19,250 frames, and the "slow dataset" size is 20,304 frames, with a 20 Hz frame rate. Both these sets are collected by the teacher-agent driving. From these two sets of recordings, single and multi-frame datasets are created. In the latter, each data point consists of three frames matched with the steering command recorded simultaneously with the third frame. A sampling scheme is applied to avoid neighboring, overlapping data points falling into train and validation sets while at the same time allowing both sets to contain data points throughout the recording duration.
    \item In the study of counteracting delays by shifting labels in the training data, we create data sets in which the captured camera frames are matched with either the simultaneously recorded command or with a command recorded up to 100 ms before and up to 200ms after the frame capture (with a step of 50ms, due to 20Hz recordings). We call these \emph{label-shifted} data sets, based on which label-shifted models are created. The underlying data is recorded by a proficient human expert at a very high speed of an average of 8.33 seconds per lap, with a standard deviation of 0.414. The data is divided into training and validation sets with a random 80/20 percent split (46620 and 11655 frames, respectively). Off-policy evaluation was performed on a separate test set of approximately 46000 frames.
\end{itemize}

\subsection{Architectures and training}

For the study of speed, four types of models were trained:
\begin{enumerate}
    \item Single-frame architecture, trained on fast data.
\item Single-frame architecture, trained on slow data.
\item Multi-frame architecture, trained on fast data.
\item Multi-frame architecture, trained on slow data. 
\end{enumerate}

For single-frame models, five-fold cross-validation was performed by dividing the data into 5 blocks along time dimension. Splitting by random shuffle would have resulted in very similar data points, captured 50 ms apart, ending up in train and validation sets. In off-policy evaluations, the averages across the five models are reported. For on-policy evaluation, a new model was trained on the entirety of the given-speed dataset to make maximal use of the data and achieve the best possible performance.

For multi-frame models, each sample consists of 3 frames and a label. Overlapping sets of frames may not end up in training and validation sets due to the high correlation between neighboring samples, resulting in inflated off-policy metrics. At the same time, we wish not to simply split the entire dataset into five pieces by time as driver ability and lighting conditions may change over time. As a solution, we first split the dataset into a number of periods along the time axis and assign 1/5 of each period into each fold. This way, each fold contains minimally overlapping pieces from all over the dataset. Again, for each speed, averages of 5 models of the cross-validation are reported in off-policy evaluation, whereas a sixth model is trained on the entire dataset for on-policy evaluation.

The architectures used are summarized in Tables \ref{tab:architecture_single} and \ref{tab:architecture_multi}.

\begin{table}[h]
    \centering
    \begin{tabular}{c|l|c|c}
Layer	& Hyperparameters	& Dropout &	Activation\\
\hline
Input & shape (height,160,3) & none & none\\
Conv2d	& filters=24, size=5, stride=2&	0.2& ReLU\\
Conv2d	&filters=32, size=5, stride=2	&0.2&	ReLU\\
Conv2d	&filters=64, size=5, stride=2	&0.2&	ReLU\\
Conv2d	&filters=64, size=3, stride=1	&0.2&	ReLU\\
Conv2d	& filters=64, size=3, stride=1&	0.2&	ReLU\\
Flatten	& -& -&	-	\\
Linear	&nodes=100	&0.2&	ReLU\\
Linear	&nodes=50	&0.2&	ReLU\\
Linear	&nodes=1	&none&	none
    \end{tabular}
    \vspace{0.2cm} 
    \caption{\textbf{Single-frame lateral-control neural network model used in this work.} This is the default architecture of the Donkey Car open-source software. Padding was always set to "valid". This architecture was used for single-frame models in the study of speed, as well as the only model type in the study of delays. \textit{height} is 60 pixels in the study of speed, and 120 pixels in the study of delays, as only the former uses cropping.}
    \label{tab:architecture_single}
\end{table}

\begin{table}[h]
    \centering
    \begin{tabular}{c|c|c|c}
Layer	& Hyperparameters	& Dropout & Activation\\
\hline
Input & size= (3, 60,160,3) & -  &- \\
Conv3d	& filters=16, size=(3,3,3), stride=(1,3,3) & -& ReLU\\
MaxPooling3D & pool\_size=(1,2,2), stride=(1,2,2) & -& -\\
Conv3d	&filters=32, size=(1,3,3),  stride=(1,1,1) &- & ReLU\\
MaxPooling3D & pool\_size=(1,2,2), stride=(1,2,2) & -& -\\
Conv3d	&filters=32, size=(1,3,3), stride=(1,1,1) & -& ReLU\\
MaxPooling3D & pool\_size=(1,2,2), stride=(1,2,2) & -& -\\
Flatten	& - & - &	-	\\
Linear	& nodes=128, batch normalization & 0.2 & ReLU\\
Linear	&nodes=256, batch normalization & 0.2 &	ReLU\\
Linear	&nodes=1& -	&none
    \end{tabular}
    \vspace{0.2cm}
    \caption{\textbf{Multi-frame lateral-control neural network model used in the study of speed.} Padding is set to "valid" in all cases.}
    \label{tab:architecture_multi}
\end{table}

The default training options in the Donkey Car repository were used. We used the mean squared error (MSE) loss function and Adam optimizer with weight decay with default parameters. Early stopping was evoked if no improvement in the validation set was achieved in 5 consecutive epochs, with the maximum epoch count fixed to 100.

\subsection{Evaluation metrics in the study of speed}
In the study of speed, the models were evaluated on-policy and off-policy. Off-policy metrics were computed using left-out validation recordings generated by the teacher agent. Mean absolute error (MAE) between model-predicted and teacher-produced values was used.

On-policy behavior was observed when deploying the models on the vehicle and letting them complete 10 laps on the track at certain speeds. The main metric was the number of infractions, i.e. collisions with walls. In case the vehicle got stuck after an infraction, it was simply placed back in the middle of the track.

\subsubsection{Out-of-distribution analysis}
The average mean squared difference between the pixels of consecutive frames is 22.4 in the slow data and 38.2 in the fast data. For multi-frame models, this means the three input images are increasingly dissimilar as speed grows. Here, we want to demonstrate that this change in inputs has a detectable effect on multi-frame models by itself, independently of the task shift of needing to turn earlier and/or sharper at higher speeds. Consequently, we can argue that the eventual driving failures at novel speeds may partly be caused by the input shift.

Networks are most accurate on their training data, as a successful optimization must by definition achieve good results on these data points. In the space of all possible data points, moving further away from the training data will likely reduce the model's performance; the speed of this drop depends on the generalization ability. For high-dimensional and complex inputs such as images, it is difficult to define what is near and what is far from training data, as pixel-wise differences do not capture the change in the meaning. Our above observation about pixel-wise differences growing with speed may or may not have a meaningful impact on our models. However, prior works have shown that out-of-distribution inputs cause detectably different activation patterns (i.e. embeddings) in the hidden layers of a network \cite{lee2018simple,hendrycks2018deep}. We wish to show that a similar change in activation patterns happens in our multi-frame models for novel-speed data. This could be interpreted as the model perceiving this data as out of distribution. Technically, we wish to show that as the speed changes, the activation patterns in the network become increasingly distinct from the patterns generated by the training data. 

To this end, for every multi-frame model in our 5-fold cross-validation, for both speeds:
\begin{enumerate}
    \item Using training data, we compute the second to last layer (final embedding layer) neuron activations, as measured in one of three possible locations on the computational graph: a) immediately after the fully connected layer), b) after applying batch normalization (BN), and c) after applying both BN and ReLU activation. For each possible extraction location, the analysis is run separately. We call these activation vectors (256 values) the \textit{reference activations}.
    \item We calculate activations also on the validation set data points. The results are referred to as \textit{same-speed activations}, as training and validation sets of any model originate from the same speed data. 
    \item Every validation sample is consequently described by a measure of \textit{distance to the reference set}, defined as the average distance to the 5 nearest reference set activations. We use Euclidean and cosine distances as proximity measures, performing separate analyses for each. \footnote{\cite{lee2018simple} proposed to use Mahalanobis distance, but our experience shows a competitive performance across different datasets with these simpler metrics.}
    \item We calculate the activation patterns using the same model for the entirety of the other-speed dataset. We call these the \textit{out-of-distribution activations} or \textit{novel speed activations}. We compute the distance of these activation patterns to the reference set according to the same metric, the average distance to the 5 most similar reference activations.
    \item Approximately, the further the activation patterns are from training patterns, the more out-of-distribution the data point is judged to be for the given model \cite{lee2018simple}. By setting a threshold on this distance, we can attempt to separate the same speed and novel speed activations. The AUROC of such a classifier is computed and presented as the main separability metric. We also report the average distances to the reference activations for same-speed and OOD-speed activations.
\end{enumerate}

These steps presented as a pseudo-algorithm are given in the Appendix.

\subsection{Setup of the study of counteracting computational delays}
In the second study, we focused on the effect of computational delays on the driving ability of good imitative models. This effect can be measured only at deployment, delays do not play a role when computing off-policy metrics. Notice that this study is related to the study of the effects of speed. In terms of distance, the effects of higher speed and higher computational delays are similar. Decisions become less spatially frequent, and the distance traveled while processing a frame increases. 

Here, we consider the main source of computational delays to be the time it takes for the neural networks to transform a camera image into a steering command. Our mini-car and software are set up to process one (most recent) frame at a time. Hence, our computational delay simultaneously increases the delay between observation and action and reduces decision frequency. In another setup, processing of multiple frames simultaneously might be possible (e.g. uploading frames to a server, computing each on a different node) and decision frequency could be maintained even when compute time and delay increase. 

In the studied case, we can artificially increase the computational delay by asking the model to wait a certain amount of milliseconds before sending out the command and processing another frame. This imitates deploying the model in a weaker computing environment or using a larger neural network model. With no artificially added delay, the average computing delay is around 24 milliseconds (99-th percentile is 40 ms), so the usual solution of behavioral cloning which learns to imitate commands captured simultaneously with the frame, deploys actions on average 24ms late (if we only consider delay from the neural network computation). 

As a first step, we investigate how the performance deteriorates if we had a similarly capable predictive model but computing slower. This would be the case of using an overly complicated model, running other systems in parallel to the steering model, or even just a sub-optimal software setup. For this, we just add artificial delay, in 25ms increments, to the computing pipeline. We deploy these pipelines and measure on-policy performance.

Secondly, we train models on label-shifted datasets. While it is possible to shift labels so that frames correspond to actions recorded prior to frame capture, this only makes decisions more belated (evaluation for this case is given in the Appendix). The primary object of study is models trained to match frames with actions recorded later than the frame capture, to predict commands that are relevant in the future. We deploy these models on the track and again measure on-policy performance.

\subsection{Evaluation metrics in the study of delays}
As discussed above, higher speeds and higher delays have similar effects on driving. Consequently, driving slowly can compensate for high computational delays. While we wish to know if our models are, in principle, capable of driving at any speed, the main goal is still to drive in the range of speeds the data was collected with, i.e. perform the designated task. 

Our metrics rely on determining the minimal lap time the model is capable of achieving in different delay conditions. Each model was deployed, and the speed was gradually increased over many laps until the vehicle started to crash regularly. The speed was then readjusted to just a fraction slower, and the vehicle attempted around 25 laps at this highest speed. At which speed to perform the 25 laps was a decision of the human experimenter, judging by the number of crashes and the overall performance on the track. Sometimes the experiment was repeated, if the experimenter doubted his decision for the speed to use or if lighting conditions seemed to disturb the performance. The average lap time of successful laps at this highest possible speed was measured from recordings and is reported. 

Based on these fastest lap times, we first show which models at which computational delays are capable of performing the trained task, defined as driving safely not slower than the training set mean speed + 2 standard deviations ($8.33+ 2\cdot 0.41$ seconds). Beyond this binary evaluation, we also report the numerical average values of these fastest safely completed laps for all model-delay combinations. Due to the combined effects of speed and delay, we can assume that a model able to drive at a faster speed has been less affected by delays. Whether label-shifting counteracts the effect of delays would be visible in these lap times.

\section{Results}

\subsection{Changing speed causes a shift in the task}
We evaluated four types of models (slow/fast training data; single/multi-frame architectures) using on-policy and off-policy measures. All models were evaluated in the conditions of slow and fast driving, i.e. also in the speed condition not trained for. 

Off-policy evaluation was performed via 5-fold cross-validation for same-speed data and on the entire dataset for novel-speed data. The results are presented in Table \ref{tab:offpol}. In short, on average the models perform better at speeds they are trained at, as expected. When transferring models trained on fast speeds to slower data, the decrease in performance is less pronounced. This is because all our models tend to underestimate steering values, and this, by chance, aligns well with the slower driving needing less extreme steering.

\begin{table}[h]
    \centering
    \begin{tabular}{|l|c|c|}
    \hline
    model type & validation data speed & mean absolute error\\
    \hline
    \textbf{slow} single frame & \textbf{slow} & 0.0232\\
    slow single frame & fast & 0.0473\\
    \textbf{fast} single frame & \textbf{fast} &0.0237\\
    fast single frame & slow & 0.0266\\
    \hline
    on average: & known &0.0235\\
             & novel & 0.0367\\
    \hline
    \hline
    \textbf{slow} multi- frame & \textbf{slow} & 0.0888\\
    slow multi-frame & fast & 0.1298\\
    \textbf{fast} multi-frame & \textbf{fast} &0.0612\\
    fast multi-frame & slow & 0.0614\\
    \hline
    on average: & known &0.0754\\
             & novel & 0.0947\\
    \hline
    \end{tabular}
    \vspace{0.2cm}
    \caption{ \textbf{Off-policy evaluation.} Mean absolute error (MAE) metric computed for different architectures trained on different data, using two different types of validation data - originating from fast or slow driving. }
    \label{tab:offpol}
\end{table}

\begin{figure}[h]
    \centering
    \includegraphics[width=5cm]{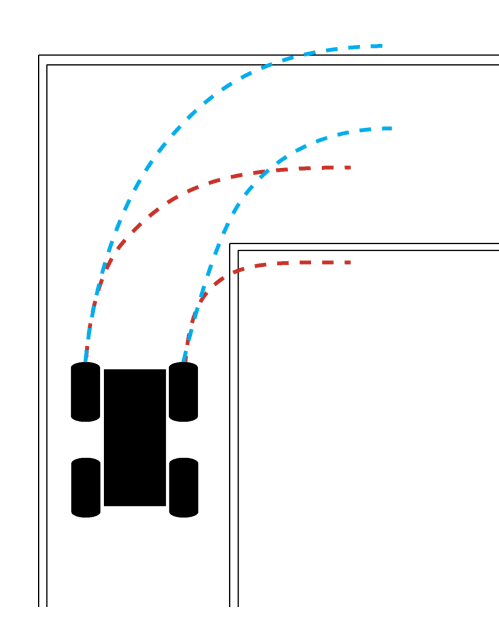}
    \caption{\textbf{Common mistakes for solutions deployed at novel speeds.} In blue: models trained on slow driving data fail to turn early and sharp enough when deployed at a fast speed. In red: models trained on fast driving data turn too early and sharp and hit the inside wall of the turn.}
    \label{fig:turns}
\end{figure}

We then deployed the models on the Donkey Car S1 hardware for on-policy evaluation. Having tuned the constant throttle value to match the slow/fast data collection speeds, we let the models control the mini-car on the same track where the data was collected. We measured the number of infractions (collisions) over 10 laps. The results are presented in Table \ref{tab:onpol} and clearly demonstrate that all models perform remarkably better at the speed they are trained on. In particular, fast-data models deployed slowly tend to cut into the inside corner of turns, while slow-data models deployed fast turn too late and end up at the outside wall (see Figure \ref{fig:turns}) 

\begin{table}[h]
    \centering
    \begin{tabular}{|l|c|c|}
    \hline
    model type & deployment speed & infractions\\
    \hline
    \textbf{slow} single frame & \textbf{slow} & 0\\
    slow single frame & fast & 10\\
    \textbf{fast} single frame & \textbf{fast} & 2\\
    fast single frame & slow & 16\\
    \hline
    \hline
    \textbf{slow} multi- frame & \textbf{slow} & 0\\
    slow multi-frame & fast & 20\\
    \textbf{fast} multi-frame & \textbf{fast} & 8\\
    fast multi-frame & slow & 19\\
    \hline
    \end{tabular}
    \vspace{0.2cm}
    \caption{\textbf{On-policy evaluation.} Infractions-per-ten-laps metric observed for different architectures trained on different data, deployed on the minicar using two different speeds - fast and slow driving.}
    \label{tab:onpol}
\end{table}

In conclusion, despite slow driving being perceived as an easier task, across two different architectures (single and multi-frame), we see that models trained to drive fast cannot reliably perform it. Not being able to perform a simpler task supports our theoretical finding about the existence of a task shift between slow and fast driving, presented in the Introduction. Notice that individual frames are very similar at different speeds (cf. Appendix), so for single-frame models, the inputs are not out-of-distribution; only the task is.

\subsubsection{Additional cause: multi-frame inputs become out-of-distribution}

For the previous results, we trained 10 multi-frame models (at 2 different speeds, 5 folds). Here, we wish to show that slow-driving data is out-of-distribution (OOD) for the model trained on fast data and vice versa. For this, we fed training data, same-speed validation data, and novel-speed validation data into each of the 10 models and extracted activation patterns from 3 possible locations along the computational graph. We used cosine and Euclidean distances as proximity metrics and computed the distance to the reference activation patterns (patterns computed on the training set) for every same and novel speed validation set activation. The average distances to the reference set (averaged over 5 models of cross-validation and the three possible extraction locations) are presented in Table \ref{tab:distances} and seem to suggest that novel-speed activations are more out-of-distribution in general. Also, it seems fast data is especially problematic for slow-data-trained models.

\begin{table}[h]
    \centering
    \begin{tabular}{|c|c|c|c|}
         \hline
         & & \multicolumn{2}{c|}{Validation data}\\
         Training data & Metric & Same speed & Novel speed\\
         \hline
         Slow  & Euclidean & 0.81 & 1.82\\
         Slow & Cosine & 0.004& 0.019\\
         Fast & Euclidean & 0.81 & 1.21\\
         Fast & Cosine & 0.006 & 0.013\\
         \hline
    \end{tabular}
    \vspace{0.2cm}
    \caption{\textbf{Multi-frame novel-speed inputs are out-of-distribution}. Average five-nearest-neighbor distances to the reference set activation patterns are presented. In all cases, novel speed causes activations clearly more dissimilar to the reference activations.}
    \label{tab:distances}
\end{table}

Further, we studied if it is possible to separate known speed (in-distribution) and novel speed (OOD) data points based on this distance. For this, a distance threshold can be set, above which all data points are classified as OOD. The performance of a such classifier across all possible thresholds can be measured via the area under the receiver operating characteristic curve (AUROC). Euclidean distance applied on activation vectors extracted after ReLU activation proved the best combination, giving an average AUROC of 0.83 for the 5 models trained on fast data and an average AUROC of 0.91 for models trained on slow data. However, for all combinations of model, metrics, and activation extraction points the AUROC values remained between 0.718 and 0.932 and showed in all cases a significantly non-trivial separability. In other words, in all cases, the activations inside the multi-frame network caused by novel-speed data could be identified with above-random performance.

We believe this demonstrates that the activation patterns caused by novel-speed data are different and potentially cause a generalization challenge for the self-driving model. Inputs being OOD does not necessarily result in lower performance if the generalization ability is high, but it is probable that the network becomes less effective in the regions of the activation space that it has not seen during training. Both this effect and task shift contribute to reduced off-policy and on-policy performance at novel speed.

We also synthesized artificial "faster" data by skipping every other frame in the slow data, so that individual images, as well as steering angles, would be identical for the slow and artificial-fast data. This means there is no additional motion blur at a higher speed, only increased inter-frame difference. The results remain qualitatively the same - larger inter-frame differences are enough to result in different activation patterns in the final embedding layer of our multi-frame models.

\subsection{Study on counteracting the effect of delays}

As described in the Methods, we trained a set of single-frame steering models, each learning to match training dataset frames with a different set of labels. These labels were obtained by shifting the rows of the labels file, to match the frame at T with the label at T + shift. With a recording frequency of 20 Hz, shifting by one row corresponds to a 50 ms shift in time. We then deployed these models and measured their on-policy performance in the presence of different amounts of computational delay.

We first asked in which conditions the solutions were capable of performing the task defined by the training set - driving at a speed similar to the training data. We notice that the baseline model with no label-shifting can perform the task at computational delays 
of 24 ms (the actual compute time) and 49 ms, but not at delays of 74 ms and above (see Table \ref{tab:shifts_binary}. This means if we had implemented a system with similar driving ability but three times slower (e.g. larger network), the model would fail to drive safely at a speed similar to the training set. For example, on our Raspberry Pi 4b device a system performing openCV-based image pre-processing steps (pinhole correction, adaptive histogram equalization) before applying the convolutional network, also incurs delays above 74 ms. Using a less downsampled input image (e.g. 4 times more pixels) would have the same effect. Clearly, the situations where the original non-shifted model would fail are realistic and could occur. 

\begin{table}[h]
    \centering
\begin{tabular}{|c||c||c|c|c|c|}
\hline
Computation time & \multicolumn{5}{c|}{Used label}\\
\hline
& No shift & 50 ms & 100 ms & 150 ms & 200 ms\\
\hline
24 ms & \textbf{+} &\textbf{+} &  &  &\\
49 ms &  \textbf{+} &\textbf{+} &\textbf{+} & & \\
74 ms &  & \textbf{+} &\textbf{+} &  & \\
99 ms &   &  &  & \textbf{+} & \\
124 ms &  &  &  & \textbf{+} & \\
\hline
\end{tabular}
\vspace{0.2cm}
\caption{ \textbf{Delay and labeling combinations that allow performing the task defined by the training set.}  The combinations allowing to drive at training set speeds or higher are marked with +. The average neural network compute delay without any modifications is 24 ms, every consecutive row adds 25 ms additional waiting time to this. The leftmost column corresponds to a model training using labels recorded simultaneously with the camera frame, other columns correspond to models trained to predict steering labels recorded after the frame capture.}
    \label{tab:shifts_binary}
\end{table}

However, the models trained to predict future commands demonstrate an increased resilience to computational delays, with the exception of the model trained with 200ms-shifted labels (discussed separately below). For example, the model trained to guess human actions 50 ms after the frame capture, can complete the task in the computational delays range of 24-74 ms. The model predicting human commands bound to occur 100 ms after frame capture can complete the task in the range of delays 49-99 ms. It is crucial to consider here that in our experimental design, the additional computational delay also reduces decision-making frequency, making completing the task harder even if all outputs are adequate. It is therefore somewhat understandable that the range of delays the 100 ms shifted model can perform at is not centered around a computational delay of 100 ms, but lower. Similarly, the model trained with a 200 ms label shift would yield timely predictions for a delay setting resulting in 5 Hz decision-making, which we know is not sufficient. Furthermore, the imitation learning task may become more complicated as we attempt to predict human actions further in time from the frame capture, so predicting commands likely to happen in 200 ms might not be as easy as the baseline task.

Based on this binary information of being able to drive at training set speed or not, we can already conclude that label-shifting successfully counteracted the effects of delays. It would allow slow-to-compute models and low-compute-power environments to complete the task where the usual behavioral cloning approach fails. This benefit costs nothing additional in terms of computing resources at deployment time.

Beyond the completion of the task, we can also see exactly how fast each model was able to drive in each condition. Here, we again make use of the fact that speed and delays have multiplicative effects - driving slower counteracts the effect of delays. Conversely, one could hence say that a model capable of driving faster has been more successful in dealing with the effect of delays.

Table \ref{tab:shifts} shows the fastest lap times in different delay and model combinations. These lap times correspond to the average over successfully completed laps at a maximal speed where the model was still able to perform, as judged by the human experimenter. The overall fastest driving was achieved with the model predicting labels 50 ms into the future in the condition of minimal possible computational delay (24 ms). If other delays (frame capture and transfer, actuator delays, etc.) were trivial, this model would deploy actions 26 milliseconds before a human would perform them. Either this preemptive behavior or simply performing human actions more reliably than the human himself, allows this model to drive faster than the fastest laps in the training dataset.

Beyond this remarkable result, the results in Table \ref{tab:shifts} show for all models the need to decrease speed when delay increases, as expected by the theoretical discussion. Looking along the columns axis, models predicting future commands tend to allow higher driving speed at similar delays, but only up to a certain time shift. A small label-shift (in our case 50 ms) seems to never hurt performance.

\begin{table}[h]
    \centering
\begin{tabular}{|c||c||c|c|c|c|}
\hline
Computation time & \multicolumn{5}{c|}{Used label}\\
\hline
& No shift & 50 ms & 100 ms & 150 ms & 200 ms\\
\hline
24 ms & \textbf{8.5} &\textbf{7.4} & $\infty$ & $\infty$ &$\infty$\\
49 ms &  \textbf{8.6} &\textbf{7.8} &\textbf{ 8} &$\infty$ & $\infty$\\
74 ms &   10.1 & \textbf{9.1} &\textbf{8.5} & $\infty$ & $\infty$\\
99 ms &  11.5 & 10.5 & 9.4 & \textbf{8.1} & $\infty$\\
124 ms & 12.5 & 11.3 & 10.6 & \textbf{9.1} & $\infty$\\
\hline
    
\end{tabular}

\vspace{0.2cm}
\caption{\textbf{Fastest lap times at different delay and label-shifting configurations.} For each model, the deployment speed is increased every few laps until the vehicle is no longer capable of driving safely. Speed just below that threshold is used to drive 25 laps.  Lap times of successful laps are determined based on recordings and the average is reported. In the table, infinity means the model can not drive a collision-free lap at any speed. Results falling into the range of training set mean +- 2 standard deviations are marked in bold.}
    \label{tab:shifts}
\end{table}

We also tested the degrading effect of predicting past labels, making the driving even more belated and requiring even lower speeds; the results of this experiment are provided in the Appendix.

\section{Discussion}

In this work, we used 1:10 scale minicars with end-to-end steering networks to prove two hypotheses about speed and delays. First, we demonstrated that deploying end-to-end steering models at novel speeds (including slower speeds) reduces their performance. In on-policy tests, all models performed clearly better on the speed their training data originated from. We believe this is at least partially due to a task shift - the function between camera frames and human-recorded steering labels is different at different speeds. To shed light on the problem, we showed that models produce on average larger mistakes on novel-speed validation data also in off-policy evaluation, not only bad performance during driving. We also showed that for multi-frame models, novel speed data has different frame-per-frame differences and causes detectably different activation patterns within the network than observations at the speed the model was trained on. We conclude that the difference between the tasks of driving fast and driving slow is significant and can influence experiment results, and the community should be aware of this fact. 

The naive solution to combat this sensitivity to speed would be to collect data at different speed ranges, but it is not always possible to collect slow-driving data. It is more time-consuming and in some countries illegal to drive unjustifiably slowly on real roads and disturb traffic. Moreover, for single-frame models, the multiple tasks (multiple speeds) contained in the training data would not be distinguishable (inputs look the same) for the model. In such cases, the model must resort to predicting the average behavior (across tasks, i.e. speeds). The alternatives are to use multi-frame models exposed to various speeds (the model can deduct speed from inputs, if necessary); deploy the model at the speed profile the data collection was performed with (might be unsafe), or use end-to-mid approaches that have formulated the self-driving problem differently and do not suffer from the speed-related task shift. Nevertheless, even for end-to-mid systems, it may prove to be useful to pay attention to the compute time during deployment and correct for the ego-motion during the computations.

Our second hypothesis was that the effects of computational delays can be counteracted by optimizing the model to map observations to future commands, e.g. the command the expert human would perform at the moment computation ends. Indeed, the behavioral cloning model that was taught to predict human commands recorded 50 ms after the camera frame capture showed better results than the baseline model optimized to pair simultaneously captured frame and command. The label-shifted model allowed faster driving and driving in the presence of larger computational delays. In general, predicting actions further into the future from observation allowed models to perform well in the presence of increased computational delays. To our knowledge, this idea to take into consideration the delays bound to occur during deployment by predicting time-shifted target labels is novel or at least not widely adopted. 

Label-shifting is cost-free in terms of additional computation cost at deployment time and counteracts the negative effect of delays independently of other remedies such as decreasing compute time or vehicle speed.

In modular and end-to-mid approaches there are more options where and how to correct the effects of delays. We believe the idea of label-shifting is relevant and applicable also in certain end-to-mid approaches. Whether it is the most effective way, remains unclear. When the steering is achieved by a control algorithm attempting to follow a neural-network-predicted trajectory in an ego-centric coordinate system, a computational delay means that by the time the target trajectory is produced, the car has already moved ahead. Either the ego-car starting position on this trajectory should be updated to include this motion (likely the easier option) or a trajectory relevant at the moment computation ends could be produced. The latter can be achieved via label shifting. Similarly, cost maps may be produced in ego-centric coordinates and may suffer from being outdated by the time computation ends. Operating in world coordinates, instead of ego-centric, would also help, but requires the existence of a localization module. In any case, being aware of the delays bound to happen and attempting to correct their effects, instead of just minimizing them, could always be attempted.

\subsection{Limitations and future work}
As an important limitation of the study of delays, we did not consider that many processes in the frame-to-action pipeline can be parallelized and a high decision-making frequency can in some cases be maintained even in the presence of noticeable delays. In our experiments, the effects of belatedness of actions and reduction of decision frequency were confounded. Our formulation reflects the problem we actually face in both our minicars and real-sized vehicle \footnote{https://adl.cs.ut.ee/lab/vehicle}, as the delay mainly comes from the neural-network computation time which can not be parallelized easily (without collecting frames into a batch, i.e. increasing delay for earlier frames). However, this formulation of the problem, due to the confounding effect of frequency, makes it harder to conclude how effective label-shifting actually is in its true task of countering belatedness. In the future, a study could be designed where decision frequency is maintained and only delay is increased artificially to then be counteracted by label-shifting.

The relevance of this work for end-to-mid self-driving, the more popular design in the 2020s, is unclear. Nevertheless, the discussions about the differences between training setup and deployment setup, and the combined effect of delays and speed may help understand the self-driving problem better in general, providing interesting angles such as thinking about spatial belatedness instead of temporal.

\section{Aknowledgements}
This work was supported by the Estonian Research Council grant PRG1604 and industry collaboration project LLTAT21278 with Bolt Technologies.

\bibliographystyle{unsrtnat}
\bibliography{references}  






\clearpage

\setcounter{section}{0}%
\setcounter{table}{0}%
\setcounter{figure}{0}%

\section*{Appendix}

\subsection*{Similarity of frames captured at different speeds}
\begin{figure*}[h]
    \includegraphics[width=14cm]{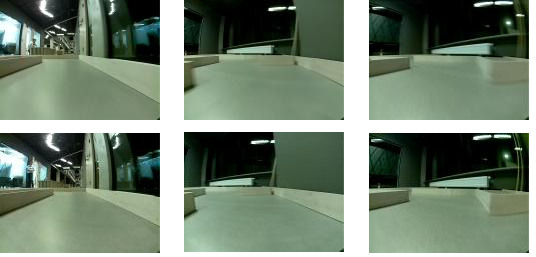}
    \caption{Frames captured at different speeds. Top row: lap speed 12 seconds per 17 meters. Bottom row: lap speed 24 seconds per 17 meters. Twice higher speed results in very slightly foggier frames during turning (center and right columns). On straight sections, the effect of speed is not perceivable in captured frames.}
    \label{fig:individual_frames}
\end{figure*}

\subsection*{AI expert driving more smooth than human}
\begin{figure}
    \centering
    \includegraphics[width=15cm]{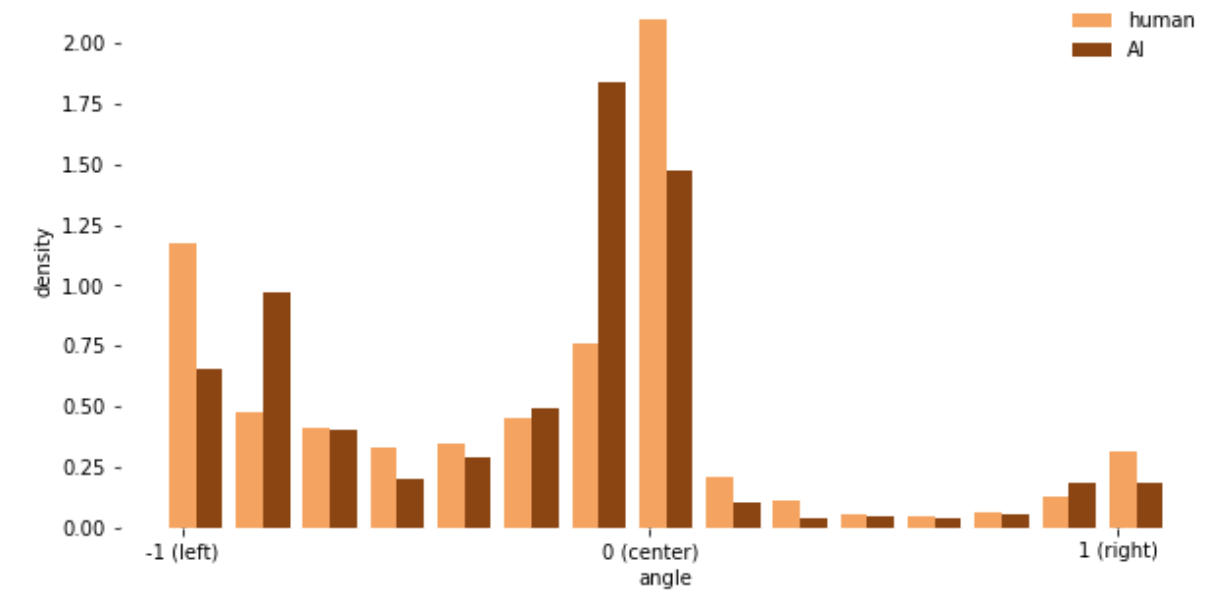}
    \caption{Human driving has more very extreme steering angles.}
    \label{fig:AI_smooth}
\end{figure}

\begin{figure}
    \centering
    \includegraphics[width=15cm]{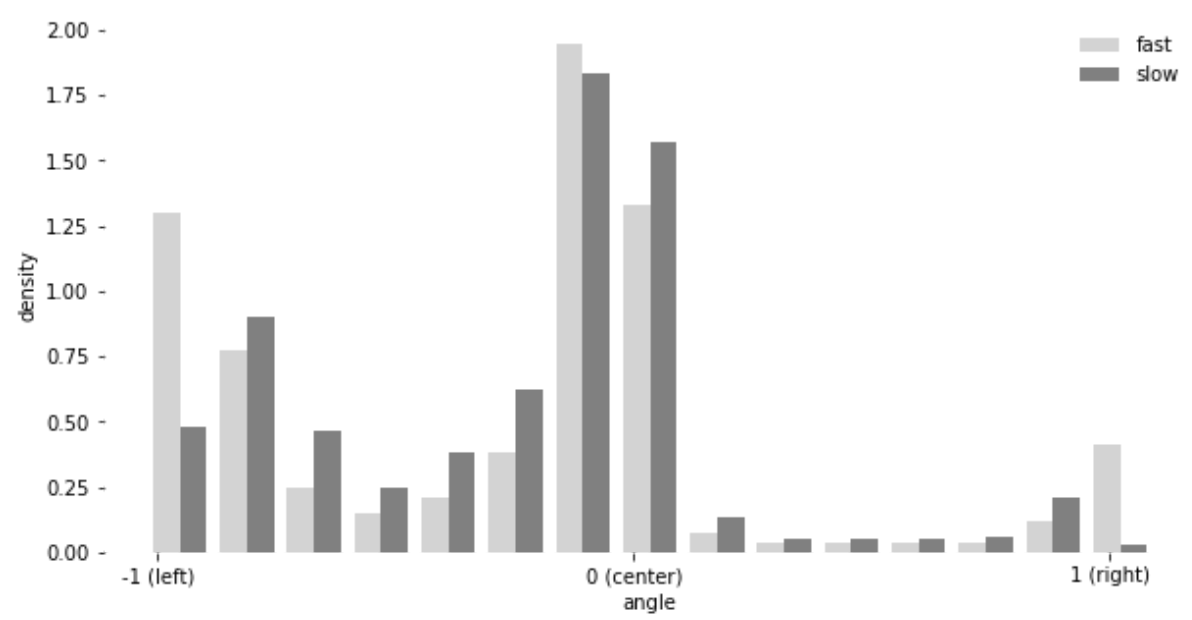}
    \caption{Even with the same AI teacher model driving fast and slow, fast speed still necessitates steeper turning.}
    \label{fig:fast_steep}
\end{figure}

\subsection*{Input becomes out-of-distribution at novel speeds for multi-frame models}
We ask the question if the activation patterns inside the network are noticeably different if we change the driving speed to something the multi-frame model has not seen during training. We choose a very simple detection method based on the five nearest neighbors' distance to activations resulting from the training set. Other OOD-detection methods could achieve a higher separation rate, in there we only wanted to demonstrate the existence of such an effect.

\subsection{Algorithm for evaluating if novel speed data is OOD}
\begin{algorithm}[H]
\vspace{1cm}
\caption{Algorithm for evaluating if novel speed causes OOD-like effects in activations}
\DontPrintSemicolon
\KwData{5 multi-frame models trained on slow data, 5 models trained on fast data, slow speed dataset, fast speed dataset}
\KwResult{Mean distance to the reference set, AUROC}
\ForEach{speed in [fast, slow]}{
    \ForEach{metric in [cosine, Euclidean]}{
        \ForEach{outputlocation in [Dense, BatchNorm, ReLU]}{
            \ForEach{fold in 5-fold cross-validation}{
                extract activations on the training set\;
                extract activations on the same speed validation set (left out fold)\;
                measure distance of validation set activations to reference set using the mean of 5 closest points according to the \emph{metric}\;
                extract activations on the novel speed dataset (OOD set)\;
                measure distance of OOD set activations to reference set using the mean of 5 closest points according to the \emph{metric}\;
                report mean distance to the reference set for the same speed validation set and for OOD set\;
                attach labels 0 and 1 to the same speed distances and novel speed distances, respectively\;
                measure how well can distance separate the 1s and 0s\;
                report AUROC\;
            }
        }
    }
}
\end{algorithm}

\subsection*{Complete table of results related to the effect of delays}
In the main manuscript, we did not cover the effect of relating recorded camera frames to the steering labels corresponding to past actions. This obviously has the effect of making steering actions increasingly belated. Lowering speed can compensate for the higher delay, as spatial belatedness equals $speed \times delay$, so such a model may still be able to complete laps at lower than training speed.
\begin{table*}[h]
    \centering
    \begin{tabular}{|c|c|c||c||c|c|c|c|}
    \hline
    Computation time & \multicolumn{7}{c|}{Used label}\\
    \hline
     & -100 ms & -50 ms & No shift & 50 ms & 100 ms & 150 ms & 200 ms\\
    \hline
    24 ms &  10.3 &\textbf{9.3} &\textbf{8.5} &\textbf{7.4} & $\infty$ & $\infty$ &$\infty$\\
    49 ms &  11.9 & 9.7 &\textbf{8.6} &\textbf{7.8} &\textbf{ 8} &$\infty$ & $\infty$\\
    74 ms &  13.2 & 11.8 & 10.1 & \textbf{9.1} &\textbf{8.5} & $\infty$ & $\infty$\\
    99 ms &  16.2 & 13.8 & 11.5 & 10.5 & \textbf{9.4} & \textbf{8.1} & $\infty$\\
    124 ms & 17.5 &13.7 & 12.5 & 11.3 & 10.6 & \textbf{9.1} & $\infty$\\
    \hline
    \end{tabular}
    \caption{Minimum safe lap time the model can consistently achieve. Intentionally predicting outdated labels recorded 50 ms and 100 ms before frame capture have been added compared to the results table in the main manuscript. The main observations still hold: a) delays can be compensated by driving slowly and b) future-predicting models allow faster driving given the same amount of delay.}
    \label{tab:shiftsfull}
\end{table*}

\end{document}